\documentclass[10pt,journal,compsoc]{IEEEtran}

\ifCLASSOPTIONcompsoc
  \usepackage[nocompress]{cite}
\else
  \usepackage{cite}
\fi
\usepackage{graphicx} 
\usepackage{subfigure}

\usepackage{epsfig}
\usepackage{amsmath}
\usepackage{amssymb}

\usepackage{algorithmic}
\usepackage{algorithm}
\usepackage{amsthm}

\theoremstyle{definition}
\newtheorem{definition}{Definition}[section]

\usepackage{algorithm}
\usepackage{algorithmic}
\usepackage{url}

\usepackage{color}
\definecolor{blue}{rgb}{0,0,1}
\definecolor{red}{rgb}{1,0,0}
\definecolor{orange}{rgb}{0.75, 0.4, 0}
\definecolor{orange}{rgb}{0, 0, 0}

\newcommand{\whitetxt}[1]{{\color{white} #1}\normalfont}
\newcommand{\rev}[1]{{\color{black} #1}\normalfont}
\newcommand{\frev}[1]{{\color{black} #1}\normalfont}

\usepackage{booktabs}
\newcommand{\ra}[1]{\renewcommand{\arraystretch}{#1}}

\usepackage[font=small]{caption}
\begin{document}

\title{Border-Peeling Clustering}

\author{Hadar~Averbuch-Elor,~\IEEEmembership{}
		Nadav~Bar,~\IEEEmembership{}
        Daniel~Cohen-Or~\IEEEmembership{}
\IEEEcompsocitemizethanks{
\IEEEcompsocthanksitem  H. Averbuch-Elor, N. Bar, and D. Cohen-Or are with Tel-Aviv University, Tel Aviv, 69978, Israel.
\IEEEcompsocthanksitem E-mail: averbuch1@mail.tau.ac.il.}
}

\markboth{IEEE TRANSACTIONS ON PATTERN ANALYSIS AND MACHINE INTELLIGENCE, JUNE 2019}%
{Shell \MakeLowercase{\textit{et al.}}: Bare Demo of IEEEtran.cls for Computer Society Journals}

\IEEEtitleabstractindextext{%
\begin{abstract}
In this paper, we present a novel non-parametric clustering technique. Our technique is based on the notion that each latent cluster is comprised of layers that surround its core, where the external layers, or border points, implicitly separate the clusters. Unlike previous techniques, such as DBSCAN, where the cores of the clusters are defined directly by their densities, here the latent cores are revealed by a progressive peeling of the border points. Analyzing the density of the local neighborhoods allows identifying the border points and associating them with points of inner layers. We show that the peeling process adapts to the local densities and characteristics to successfully separate adjacent clusters (of possibly different densities). 
We extensively tested our technique on large sets of labeled data, including high-dimensional datasets of deep features that were trained by a convolutional neural network. We show that our technique is competitive to other state-of-the-art non-parametric methods using a fixed set of parameters throughout the experiments.
\end{abstract}

}

\maketitle

\IEEEdisplaynontitleabstractindextext

\IEEEpeerreviewmaketitle

\IEEEraisesectionheading{\section{Introduction}\label{sec:introduction}}

\IEEEPARstart{C}{lustering} is the task of categorizing data points into groups, or clusters, with each cluster representing a different characteristic or similarity between the data points. Clustering is a fundamental data analysis tool, and as such has abundant applications in different fields of science and is especially essential in an unsupervised learning scenario. Ideally, a clustering method should infer the structure of the data, e.g., the number of clusters, without any manual supervision. 

Many of the state-of-the-art clustering methods operate with several underlying assumptions regarding the structure of the data.
A prominent assumption is that the clusters have a single area that can be identified as the center, or the core of the cluster. For instance, K-Means \cite{macqueen1967some} operates under the assumption that there is a single cluster center according to the compactness of the data, while the Mean-Shift \cite{cheng1995mean} method defines this area as the one displaying the highest density inside the cluster. Operating under this assumption may result in overly split clusters containing several dense areas, or centers, of smaller clusters.

Density based methods like DBSCAN \cite{ester1996density} operate often under the assumptions that different clusters have similar levels of density, and that the cores of the clusters can be defined based on density reasoning. However, often, the cores of the clusters do not have a clear structural density that can be directly defined, leading to a redundant merge of adjacent clusters.

\begin{figure*}[t]
	\centering
	\includegraphics[width=1.0\textwidth]{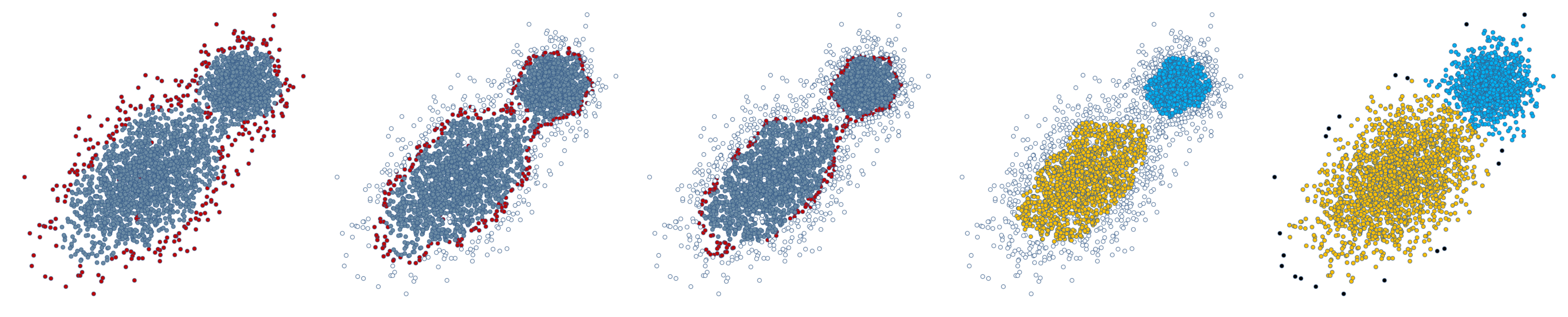}
	\includegraphics[width=1.0\textwidth]{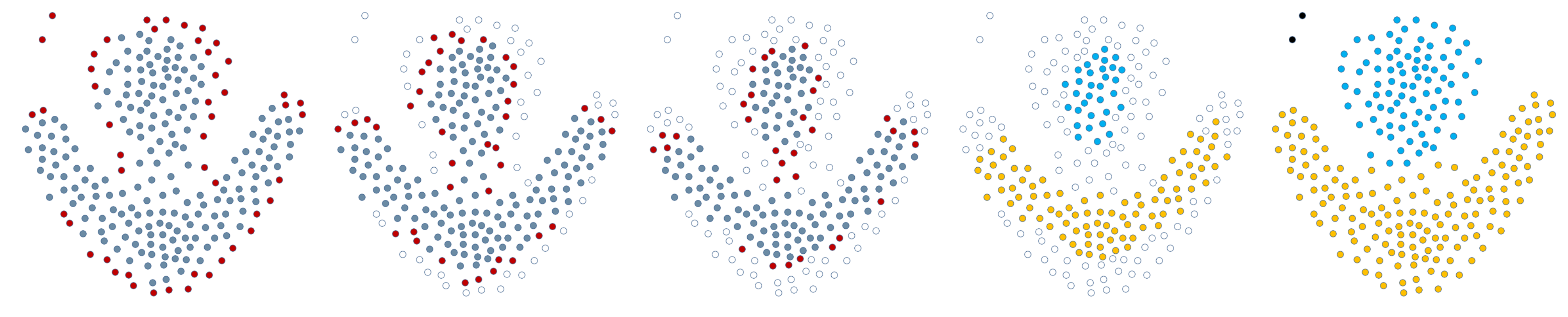}
	\caption{The Border-Peeling technique on two different datasets: Three consecutive peeling iterations are illustrated on the left, with the identified border points colored in red. Next, we illustrate the clustering of the highly-separable core points that remain after the peeling process. The rightmost figures illustrate our final clustering results (outliers are colored in black).}
	\label{fig:peeling_process}
\end{figure*}

In this work, we introduce \emph{Border-Peeling}: a non-parametric clustering method that iteratively peels off layers of points to reveal the cores of the latent clusters. The peeling is a local process, where the decisions are a result of local analysis, revealing the local behaviour of points without expecting other clusters to share similar characteristics.
Our technique is non-parametric in the sense that the number of clusters is not provided as input. The careful repeated peeling forms a transitive association between the peeled \emph{border} points and the remaining \emph{core} points. The key is to consider a layered structure for the latent clusters, where the external layers implicitly separate the clusters. We analyze the density of the local neighborhoods of each point to iteratively estimate the border points and associate them with inner-layer points. See Figure \ref{fig:peeling_process} for an illustration our iterative technique.

In the following sections, we describe the details of our method, and demonstrate the performance of our method on various datasets and in comparison to other well-known clustering methods. Except for common baseline datasets, we use large datasets, containing thousands of points, for which we have their ground truth clusters, and extensively evaluated the performance using many random subsets.
Particularly, we evaluate our technique on high-dimensional datasets of deep features that were trained by convolutional neural networks. We show that the Border-Peeling clustering algorithm \rev{is competitive to other state-of-the-art non-parametric clustering methods} using a fixed set of parameters throughout the experiments.

\section{Related Work} \label{sec:related}

Data clustering is one of the most fundamental problems in data analysis, highly applicable to various fields of science. For a comprehensive survey on data clustering techniques, please refer to \cite{jain2010data}. 
In our work, we present a novel non-parametric clustering method. Non-parametric clustering is an active field of research, receiving ongoing attention for several decades now. In what follows, we elaborate on the most closely related works.

The DBSCAN \cite{ester1996density} method groups together points that are packed together closely , marking points that lie individually in low-density regions as noise. The notion of border points within a cluster has been established by DBSCAN in the past. The border points in DBSCAN are defined as points that are part of a cluster but are not surrounded by a dense neighborhood. A notable difference between their method and ours is that in DBSCAN, the identified border points do not have a prominent role in the definition of the different clusters. Furthermore, the association to their resulting cluster is dependent on the order in which they are processed. Additionally, the DBSCAN method has been shown to be sensitive to its input parameters \cite{karypis1999chameleon}. Later works, e.g., \cite{karami2014choosing,sawant2014adaptive}, extended their work to allow for an automatic parameter estimation. 

The OPTICS \cite{ankerst1999optics} clustering technique extends DBSCAN. Unlike DBSCAN, it can detect clusters in data of varying density by producing a reachability plot with an ordering of the data points according to their clusters. However, OPTICS requires a manual analysis of the reachability plot. Furthermore, it does not have a notion of border points. HDBSCAN \cite{campello2013density}, a recent extension of both DBSCAN and OPTICS, is a hierarchical version of DBSCAN, where a flat partition, consisting of the most prominent clusters, can be extracted from the hierarchy. HDBSCAN only requires one parameter (the minimum cluster size) and can handle data of varying density. Similarly to OPTICS, HDBSCAN does not have a notion of border points. However, due to the fact that HDBSCAN chooses the density threshold automatically, by comparing hierarchies of dense areas, it tends to cluster a large number of data points as noise.

CHAMELEON  \cite{karypis1999chameleon} is a hierarchical method of clustering, which partitions the data according to its $k$-NN graph and merges the components of the graph according to their proximity and inter-connectivity. However, as discussed in \cite{ertoz2003finding}, this method may yield sub-optimal results for data in higher dimensions. Other more recent hierarchical methods have been shown to effectively handle outlier points \cite{cheng2018local,cheng2018hierarchical}. 
The work presented in \cite{ertoz2003finding} suggests a clustering technique based on a shared (mutual) nearest neighbor (SNN) approach. It has been shown that such an approach is advantageous when the goal is to find the most significant clusters, rather than identifying all the latent clusters \cite{maier2007cluster}.	

The Mean-Shift \cite{cheng1995mean} method clusters data points using a kernel density estimation function, by iteratively shifting each data point to a dense region in its proximity, and then clustering the shifted data points. It is, however, dependent on the bandwidth parameter of the kernel density estimator. The Adaptive Mean-Shift \cite{shimshoni2006adaptive} method overcomes this issue by estimating a different bandwidth for each data point according to the local neighborhood of the point. As mentioned in \cite{carreira2015review}, in many cases, Mean-Shift clustering tends to over-partition the clusters, i.e., it often returns a large number of clusters, even when the actual number of clusters is small. In our work, over-partitioning is avoided due to our peeling termination criterion, which is determined automatically according to the characteristics of the data.

The Affinity Propagation \cite{frey2007clustering} technique clusters data points according to the concept of message passing between data points. Affinity propagation performs the clustering by first finding exemplars in the dataset, and then clustering each data point according to the exemplar that the data point was associated with. Our method also uses data point association to cluster the data points. However, unlike our method, Affinity Propagation tends to over-partition the data.


The topic of identifying cluster border points was previously addressed in Xia et al.\cite{xia2006border}, where the points in the datasets are ordered according to the number of $k$-neighborhoods in which they participate (Reverse KNN), and points with a low number of neighborhoods are considered to be the border points. In their work, the border identification is presented as a general pre-processing step which may improve the clustering result. However, they do not address to the issue of classifying the points to different clusters, as well as to the clustering itself. \rev{The QCC algorithm \cite{huang2017qcc} bears some similarity to our method as it also utilizes reverse k-nearest neighbors, however, conceptually, it is quite different. Unlike our method where the cluster borders are iteratively peeled, QCC is a center-based approach, where the clustering results are driven by the estimated cluster centers.}



\section{The Algorithm}

Given a set of points in $\mathbb{R} ^d$, our clustering technique iteratively \emph{peels off} border points from the set. The premise of this method is that by repeatedly peeling off border points, the final remaining points, termed \emph{core points} will be better separated and easy to cluster. 

To cluster the input points, during the iterative peeling process, each peeled point is associated and linked with a neighboring point that was not identified as a border point. This linkage forms a transitive association between the peeled points with one of the core points. The clustering of the peeled border points is then inferred by their association to the clustered core points.

\rev{
\begin{algorithm}
\caption{Border-Peeling Clustering}
\label{alg}
\begin{algorithmic}
\rev{
\STATE \textbf{Input:} A set of points  $X=\{x_1, x_2,...,x_n\} \in \mathbb{R} ^d$. 	
\STATE \textbf{Output:} Cluster indices $C=\{c_1,c_2,...,c_n \}$.
\STATE $X^1 \leftarrow X$
\FOR{peeling iteration $ 1 \leq t \leq T$ } 
\FOR{each point $x_i \in X^t$ }
\STATE $RN_k^{(t)}(x_i) \leftarrow {\left\{x_j|x_i \in N_k^{(t)}(x_j) \right\}}$
\STATE $b^{(t)}_i \leftarrow \sum_{x_j\in RN_k^{(t)}(x_i)} \exp \left(-\frac{||x_i-x_j||_2^2}{\sigma_j^2}\right)$
\ENDFOR
\STATE $X_B^{(t)} \leftarrow \left \{ x_i  \colon B^{(t)}_i = 1 \wedge x_i \in X^{(t)} \right \}$
\STATE $X^{(t+1)} \leftarrow X^{(t)} \setminus X_B^{(t)}$
\FOR{each peeled point $x_i \in X_B^{(t)}$ }
\STATE $\rho _i \leftarrow$ \textsc{AssociatePoint}$(x_i,X^{(t+1)})$ \COMMENT{see Section \ref{sec:BorderAssoc}}
\ENDFOR

\ENDFOR
\STATE $\tilde{c} \leftarrow$ \textsc{ClusterCorePoints}$(X^{(t+1)})$ \COMMENT{see Section \ref{sec:clust}}
\STATE $c \leftarrow$ \textsc{ComputeFinalResult}$(X, \tilde{c}, \rho)$ \COMMENT{linking core point clustering to computed associations}
}
\end{algorithmic}
\label{alg1}
\end{algorithm}
}


The algorithmic key idea of our technique is twofold: (i) the definition of a border point, and (ii) the association of a border point to its neighboring non-border point. These key ideas will be elaborated in the following section.

The iterative peeling terminates when the identified border points are strictly weaker in terms of their "borderness" than the border points that were identified in the previous iterations, thus forming the set of core points. These core points are then grouped into clusters using a simplified version of DBSCAN. \rev{In Algorithm \ref{alg1}, we describe our algorithm in pseudo-code.}

In what follows, we first introduce some notations and describe how border points are identified at each iteration (Section \ref{sec:BorderIden}). We then detail the border point association process (Section \ref{sec:BorderAssoc}). Finally, we describe our clustering procedure (Section \ref{sec:clust}).

\subsection{Border Points Identification} \label{sec:BorderIden}
\rev{

Given a set of $n$ data points $X = \{x_1, x_2,...,x_n\}$ in $\mathbb{R} ^d$ and a dissimilarity function $\xi: \mathbb{R} ^d\times \mathbb{R} ^d \to \mathbb{R}$ as input, we denote $X^{(t)}$ as the set of points which remain \emph{unpeeled} by the start of the $t^{th}$ iteration.

For each point $x_i \in X^{(t)}$, denote the set of $k$ nearest neighbors by $N_k^{(t)}(x_i)$. Following \cite{korn2000influence} and \cite{xia2006border}, the reverse $k$ nearest neighbors of $x_i$ is given by
\begin{equation}
RN_k^{(t)}(x_i)={\left\{x_j|x_i \in N_k^{(t)}(x_j) \right\}}
\end{equation}
That is, $RN_k^{(t)}(x_i)$ is the set of points for which $x_i$ is one of their k-nearest neighbors.

To estimate distances between points, we use a pairwise relationship function $f:{\mathbb{R}^d \times \mathbb{R}^d  \to \mathbb{R}}$ . In our work, where the dissimilarity measure $\xi$ is the Euclidean distance, we calculate $f$ by applying a Gaussian kernel with local scaling  \cite{zelnik2004self} to the Euclidean distance:
\begin{equation}
f(x_i,x_j)=exp \left(-\frac {||x_i-x_j||_2^2}{\sigma_j^2}\right),
\end{equation}
Following the choice of $\sigma_j$ in \cite{zelnik2004self} we set  $\sigma_j=||x_j-N_k^{(t)}(x_j)[k]||_2$. where $N_k^{(t)}(x_j)$ denotes the $k$-th nearest neighbor of $x_j$ at iteration $t$, i.e. the distance to the $k^{th}$ neighbor of the data point is used as the normalizing factor for the Gaussian kernel. This approach has proven to be effective in measuring the affinity between data points when the affinity of the data points has a large variance.

Using $f$, we associate with each point in $X^{(t)}$ a \emph{density influence} value $b^{(t)}_i$ where
\begin{equation}
b^{(t)}_i=\sum_{x_j\in RN_k^{(t)}(x_i)} {f(x_i,x_j)}.
\end{equation}
The density influence $b_i$ aims at capturing the amount of influence a point has on the local density of its neighboring points. We would expect the values of $b_i$ to be smaller for data points that lie on the border of the cluster, and larger for points that are closer to the core of the cluster. In the supplementary material we analytically demonstrate $b_i$'s desired behavior for the simple case where the points are uniformly distributed over a section in $\mathbb{R}^1$.

Recall that in every \emph{peeling} iteration, the algorithm classifies some of the points of $X^{(t)}$ as border points and \emph{peels} them off. Formally, for each point $x_i\in X^{(t)}$, denote by $B^{(t)}_i$ the \emph{border classification value} of that point that accepts the value of $1$ if $x_i$ is a border point and $0$ otherwise. The calculation of $B^{(t)}_i$ in each iteration is performed using an iteration specific cut-off value $\tau^{(t)}$:
\begin{equation}
    B^{(t)}_i= 
\begin{cases}
    1,& \text{if } b^{(t)}_i \leq \tau^{(t)}\\
    0,              & \text{otherwise}
\end{cases}
\end{equation}
It is important to note that $B^{(t)}_i$ is space-variant due to its reliance on $f$ (through $b^{(t)}_i$). Simply put, we learn the local characteristics of the dataset to determine whether a point is classified as a border point or not. The set of cut-off values $\tau^{(1)},\tau^{(2)}...$ can be manually specified, or as we describe below, it can be estimated from the intrinsic properties of the input data.

To conclude the \emph{peeling} iteration, the set of border points at iteration $t$ is given by
\begin{equation}
X_B^{(t)} = \left \{ x_i \in X^{(t)} \colon B^{(t)}_i = 1\right \}
\end{equation}
and the set of \emph{unpeeled} data points for the next iteration by
\begin{equation}
X^{(t+1)}=X^{(t)} \setminus X_B^{(t)}
\end{equation}
}

\begin{figure*}[t!]
	\centering
	\centering
	\includegraphics[width=0.47\textwidth]{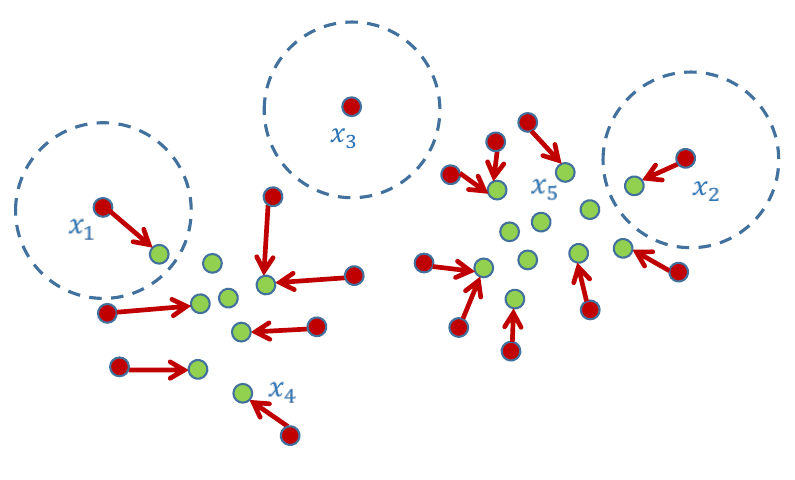}
	\hspace{15pt}
	\includegraphics[width=0.47\textwidth]{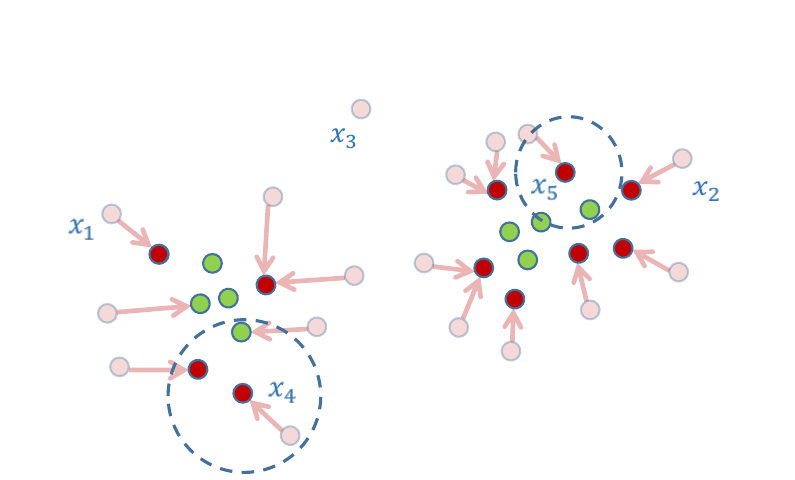}
	Iteration 1 \hspace{165pt} Iteration 2
	\caption{The border association process and calculation of $l_i$ during the peeling process. The figures illustrate the border points (shown in red) of the first two iterations and their association to non-border points, marked by the red arrows. The blue circles represent the association area induced by the spatially-variant threshold value $l_i$. 
	}
	\label{fig:association}
\end{figure*}

\subsection{Border Points Association}\label{sec:BorderAssoc}

\rev{
Following the identification of border points at iteration $t$, we associate to each identified border point $x_i \in X_{B}^{(t)}$ a neighboring non-border point which we denote as $\rho_i \in X^{(t+1)}$. In order to prevent a scenario in which an isolated border point is associated with a relatively distant neighbor, resulting with erroneous merging of distant clusters, we mark some points as outliers. These points will not be part of any cluster.

$\rho_i$ is given by
\begin{equation}
    \rho_i = 
\begin{cases}
    x_j,& \xi(x_i,x_j)\leq l_i\\
    \emptyset,& \xi(x_i,x_j)>l_i\\
\end{cases}
\end{equation}
where $x_j$ is the closest non-border point to $x_i$ at iteration $t$ from the set of non-border points and $l_i$ is a spatially-variant threshold value. $\emptyset$ is used to mark $x_i$ as an outlier. In words, if there are no non-border points within a distance of $l_i$ from $x_i$ then $x_i$ is marked as an outlier. Otherwise, $\rho_i$ is the nearest non-border point to $x_i$ at iteration $t$.

For each border point $x_i$, $l_i$ is determined at the time the point is classified as a border point. All border points identified in the first iteration receive the value of $\lambda$, where $\lambda$ is a parameter of our method which serves as the maximal threshold value. For the subsequent iterations, consider a point $x_i$ at the iteration $t$ where it was peeled. We take the $k$-nearest data points to $x_i$ from the data points that are in $\cup_{r=1}^{t}X_B{(r)}$, i.e., the set of points that were already peeled up to the current iteration and were not marked as outliers. We denote this set by $NN_{B,k}^{(t)}(x_i)$. We then compute
\begin{equation}
l_i = \min \left\{\frac{C}{k}\sum_{x_j \in NN_{B,k}^{(t)}} \xi(x_i,x_j)  , \quad \lambda \right\}
\end{equation}
where the constant $C$ determines the strictness of the threshold values ($C=3$ in all our experiments).

We have found this method to perform better than when using a constant threshold value, as it takes into account the spatially-varying density of the data points. See Figure \ref{fig:association}.

Figure \ref{fig:association} illustrates the effects of the spatially-variant threshold value $l_i$. As the figure demonstrates, at first the association areas (illustrated by blue circles), whose radius equals $l_i$, are of equal size. Some data points (such as $x_3$ in the figure) are not associated to a non-border point since there are no non-border points with distance of at most $\lambda$.  The right figure illustrates the threshold values after one iteration. \frev{The values of the newly identified border points are calculated by averaging over the Euclidean distances to the nearest peeled border points. Note, for example, that $l_4$ is larger than $l_5$, as $x_4$ is further away from its nearest peeled border points (assuming, for instance, $k=3$).}
 }

\subsection{Bottom-up Clustering} \label{sec:clust}

Our method iteratively identifies border points and associates them to non-border points. 
The transitive association thus yields paths from each input point to one of the core points, which are the final remaining non-border points, i.e., the set $X^{(T+1)}$ with the set that contains $\rho_i$. We define a heuristic to set the number of iterations $T$ in a context-sensitive manner (more details are provided in Section \ref{sec:exp}).

Once the border peeling process terminates, the remaining set of core points are clustered by merging together close reachable neighborhoods of points. 
\rev{Formally, we define reachable neighborhoods of core points as follows:

\theoremstyle{definition}
\begin{definition}{}
A pair of core points $x_i,x_j \in X^{(T+1)}$ is said to be reachable w.l.o.g. if there is a series of core points: $\left(x_{k_1},...x_{k_m} \right)$ with $k_1=i$ and $k_m=j$, such that for every two adjacent indices in the series $(k_{r},k_{r+1})$ the relation $\xi(x_{k_r},x_{k_{r+1}}) \leq max(l^{(T)}_{k_r}, l^{(T)}_{k_{r+1}})$ holds.
\end{definition}
}

For every pair of core points $(x_i, x_j)$ that are reachable, we merge the set that contains $x_i$ with the set that contains $x_j$. This merging step is done iteratively, until all sets of reachable data points are merged.

The set of cluster candidates are then defined by following the border points association and linkage to the core points. In order to be able and better filter out noise, we mark small clusters as noise, using a user defined value of the minimum cluster size. Following the filtering step, the final set of clusters is returned.

\section{Experiments} \label{sec:exp}

To evaluate the performance of the Border-Peeling clustering method, we measured its performance on numerous synthetic and real-life datasets and compared its performance to other state-of-the-art algorithms. For each experiment, the free parameters of all the alternative methods were set according to their \emph{best} performance over a large range of possible configurations. Contrarily, for our method, a \emph{fixed} set of parameters was used throughout \emph{all} of the experiments. This gives a huge advantage to the alternative methods and demonstrates that our method is insensitive to its parameters setting (which we further validate in \rev{a sensitivity evaluation which can be found in the supplementary material}). 

The free parameters of our method were set as follows: 
In each iteration, $\tau^{(t)}$ was set such that $90\%$ of the remaining data points at each iteration have larger values of $b_i^{(t)}$, and a value of $k=20$ was used for the k-NN queries.  The value of $\lambda$ was calculated by first calculating all of the pairwise distances in the k-neighborhood of each point: $D_k = \cup_{x_i \in X} \left \{ d(x_i,x_j) \middle| x_j \in N_k(x_i) \right \}$, and then setting $\lambda = MEAN(D_k) + STD(D_k)$. While other choices of $\lambda$ can be used, we have found this simple estimation method to be effective, as can be seen in the results below.

To automatically set the number of iterations of the peeling process $T$, we make the following observation: peeled border points should reside in denser areas than the border points of the previous iteration. Therefore, the assigned density influence values should increase in average (which implies that the points establish less coherent borders than the previous iterations). Furthermore, when over-clustering occurs, the values of the border points are expected to be significantly higher than the values of the border points in the preceding iterations.

%

Hence, in each iteration $t$, we track the set of values of border points that are about to be peeled: $\left \{ b_i^{(t)} \middle| B_i^{(t)} = 1\right \}$, and calculate the mean value of that set, denoted by $\bar{b}_p^{(t)}$. We then examine the the values of $\bar{b}_p^{(1)}, \bar{b}_p^{(2)},...,\bar{b}_p^{(t)}$, and terminate the peeling when $ \frac{\bar{b}_p^{(t)}}{\bar{b}_p^{(t-1)}} - \frac{\bar{b}_p^{(t-1)}}{\bar{b}_p^{(t-2)}} > \epsilon$, where $\epsilon$ is a positive constant that we empirically set to $0.15$. 

We compare our performance to that of a set of well-known non-parametric clustering techniques: DBSCAN (DB) and the more recent Hierarchical-DBSCAN (HDB) \cite{campello2013density}, Mean-Shift (MS), Affinity Propagation (AP) \cite{frey2007clustering}, QCC \cite{huang2017qcc} and Robust Continuous Clustering (RCC) \cite{shah2017robust}. Similar to Border-Peeling clustering, all of those clustering methods try to infer the number of clusters automatically and do not accept the number of clusters as a parameter. We also compare our performance to the K-Means (KM) and Spectral Clustering (SC) \cite{ng2002spectral} methods, which accept the number of clusters as a parameter.  For each dataset, we run K-Means 1000 times with random initialization, taking the clustering which minimizes the sum of distances of data points to their closest cluster center among those 1000 runs. Similarly, for Spectral Clustering, the K-Means phase is also run 1000 times, taking the best result among these runs.

In what follows, we explain and demonstrate our quantitative comparisons on three synthetic datasets as well as on sampled large datasets. In the supplementary material, we provide qualitative results on six additional synthetic datasets of various distributions and further illustrate the huge advantage we provide the alternative techniques by presenting the entire grid of results. While all our results are obtained with a fixed set of parameters, we explicitly demonstrate how for alternative methods a fixed parameter setting will work well for only some of the scenarios.  

\begin{figure}
	\centering
	A \hspace{65pt} B \hspace{65pt} C \\ 
	\vspace{10pt} 
		\rotatebox[origin=c]{90}{\whitetxt{2222222222222} K-Means} \includegraphics[width=0.45\textwidth]{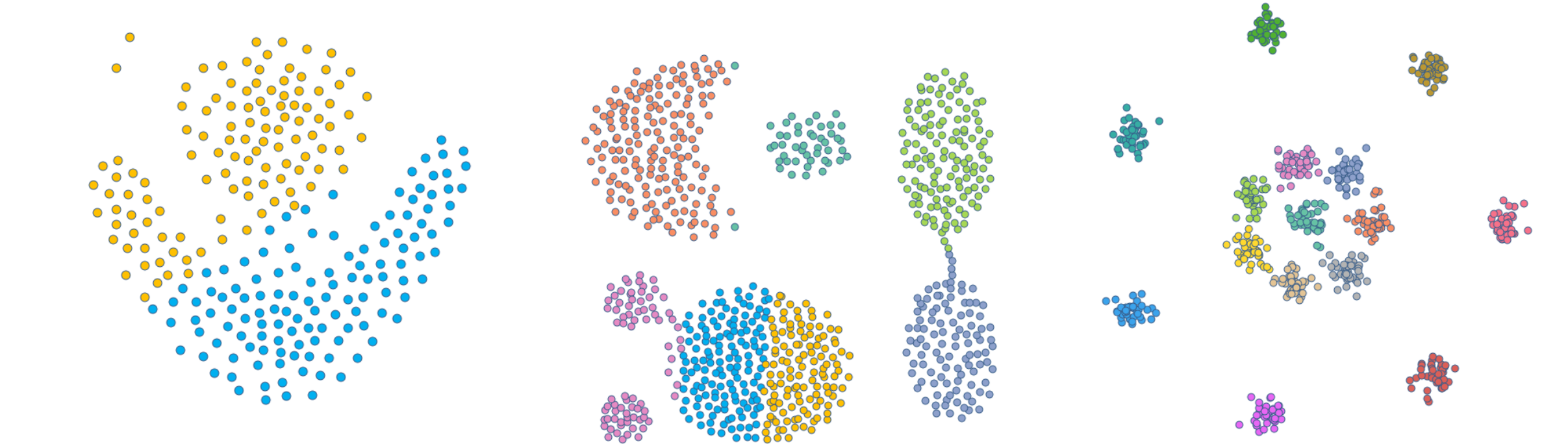} \\
	\vspace{-50pt}
	\rotatebox[origin=c]{90}{\whitetxt{22222222222222222} Spectral Clustering} \includegraphics[width=0.45\textwidth]{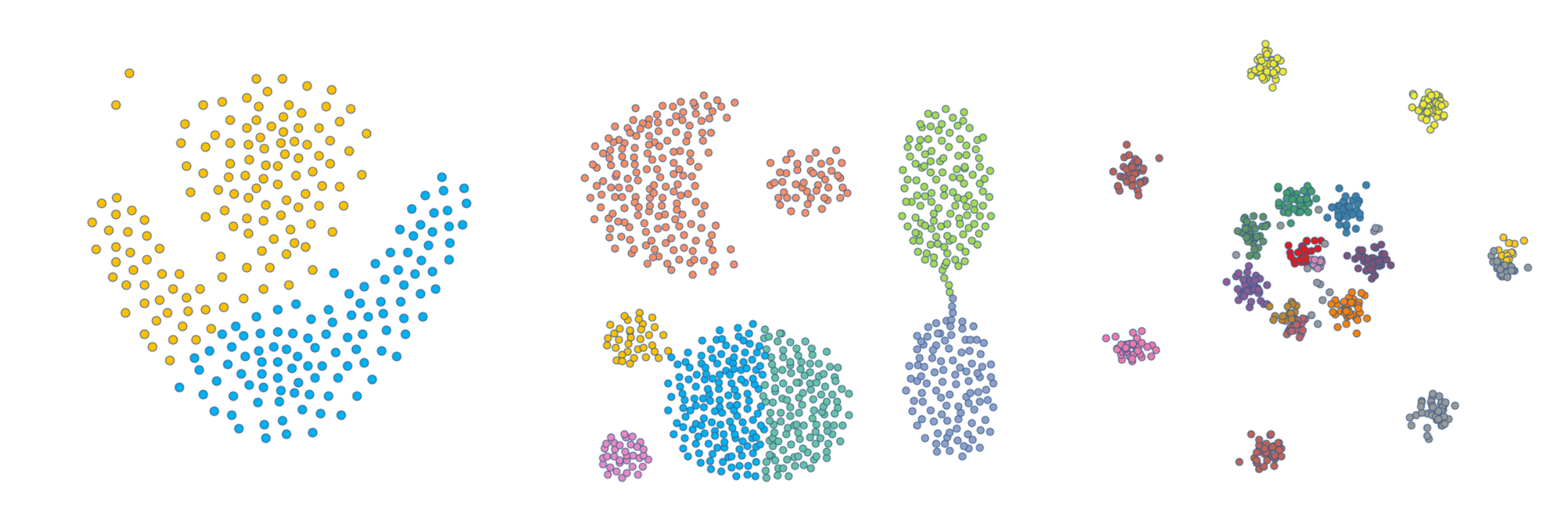} \\
	\vspace{-75pt}
	\rotatebox[origin=c]{90}{\whitetxt{2222222222222} Border-Peeling}   \includegraphics[width=0.45\textwidth]{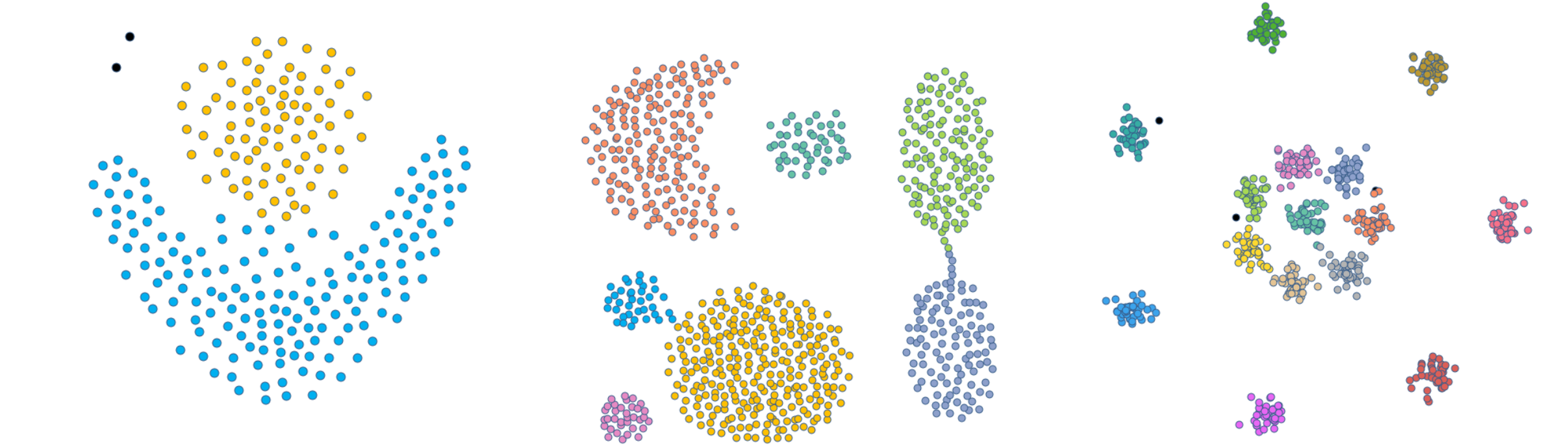}
	\vspace{-55pt}
	\caption{Qualitative comparison of our Border-Peeling technique to parametric clustering techniques on three synthetic examples from the literature (labeled A-C). As the figure demonstrates, our technique successfully recovers the number of clusters automatically as well as identified the outliers (colored in black), while the two parametric methods fail to identify the correct clusters for datasets A and B. }
		\label{fig:toy_cluster}
\end{figure}

\begin{table}
\centering
\ra{0.9}
\setlength{\tabcolsep}{2.3pt}
\begin{tabular}{@{}lllcccccccccccccc@{}}
\toprule
\multicolumn{2}{c}{} &  & \multicolumn{3}{c}{A (2 Classes)} & \phantom{}& \multicolumn{3}{c}{B (7 Classes)} & \phantom{} & \multicolumn{3}{c}{C (15 Classes)}\\
 && & ARI & AMI & \rev{Det\#} &&  ARI & AMI & \rev{Det\#} &&  ARI & AMI  & \rev{ Det\#}\\ \midrule
 KM && & 0.453 &0.386 & - && 0.526   & 0.716 & -&& \textbf{0.993} & \textbf{0.994}& - \\  
 SC && & 0.388 &0.411& - &&  0.731  & 0.857 & -&& 0.726 & 0.851& -\\ \midrule
 DB && & 0.922 &0.826& \rev{2} && 0.992   & 0.982& \rev{7} && 0.927 & 0.926& \rev{15}\\  
 HDB && & 0.754 &  0.574 & \rev{2}&& 0.809  & 0.799& \rev{5}  && 0.956 & 0.952& \rev{15}\\ 
 AP && & 0.138 &0.240& \rev{12} && 0.393  & 0.601& \rev{16} && \textbf{0.993} & \textbf{0.994}& \rev{15} \\ 
 MS && & 0.562 &0.826& \rev{2} && 0.526  & 0.716 &\rev{ 5} && 0.915 & 0.952& \rev{15}\\  
 QCC && & 0.950 &0.897& \rev{2} && \textbf{0.997}   & \textbf{0.995}& \rev{7} && 0.980 & 0.982& \rev{16}\\  
 RCC && & 0.934 &0.881 & \rev{2}&& 0.604  & 0.701& \rev{14} && 0.982 & 0.985& \rev{15} \\ 
 BP && & \textbf{0.983} &\textbf{0.962}& \rev{2} && 0.996 &0.992& \rev{7} &&  0.982 & 0.985& \rev{15} \\

\bottomrule
\end{tabular}
	\caption{Comparison of Border-Peeling clustering (BP) with parametric (KM, SC) and non-parametric (HDB, DB, AP, MS, QCC, RCC) clustering techniques on the common synthetic datasets A, B and C which are illustrated in Figure \ref{fig:toy_cluster}. \rev{Alongside the scores, we also report the total number of output clusters (Det\#) obtained for each experiment for the non-parametric techniques.}
}
\label{tab:toy}
\end{table}
\subsection{Evaluation on common synthetic datasets}

First, we quantitatively evaluated our method on a number of common synthetic datasets from the known clustering literature \cite{fu2007flame,gionis2007clustering,veenman2002maximum}, which consist of a small number of clusters that lie in proximity to one another and are not easily separable, thus constituting a challenge to density-based methods. 
We provide a comparative visualization of the clusters formed by Border-Peeling clustering to the clusters formed by K-Means and Spectral Clustering, to demonstrate that our non-parametric clustering method outperforms parametric techniques even in synthetic settings (see Figure \ref{fig:toy_cluster}). 

We evaluate the performance of each method by calculating the Adjusted Rand Index (ARI) \cite{hubert1985comparing} and Adjusted Mutual Information (AMI) \cite{vinh2010information} of the resulting metrics. The ARI and AMI are well-known metrics in the field of data clustering and are frequently used in order to evaluate clustering performance when ground truth data is available. The scores for the well-known clustering methods were obtained by running them over a range of parameters and taking the clustering result with the best AMI score, while Border-Peeling clustering was run using the parameters described in the previous section. See Table \ref{tab:toy} for the full comparison on these synthetic datasets. As Figure \ref{fig:toy_cluster} demonstrates, our technique successfully identified the number of clusters for each of the datasets and achieved the best scores overall.

\subsection{Evaluation on large datasets}

We further evaluated the performance of the Border-Peeling method in comparison to other well-known clustering algorithms by running it on large datasets. \frev{We perform two different sets of experiments: (i) experiments on \emph{unsupervised} features and (ii) experiments on \emph{supervised} (pretrained) features. As can be expected, the results are generally better on the supervised features. However, as the evaluation illustrates, clustering a large dataset is challenging even in a supervised setting (see Figure \ref{fig:mnist_cifar} for a visualization of the supervised features). Our evaluation shows that our method is competitive to other state-of-the-art techniques in both cases. It is important to note that throughout our experiments, the Border-Peeling parameters were fixed as described above, while the parameters of the other techniques were set separately for each dataset by taking the parameters that yield the best result. Furthermore, note that in this section, we only partially compare to the QCC technique as their implementation is in MATLAB, making it infeasible to run all the different configurations on these large datasets. 

}

\subsubsection{Experiments on unsupervised feature vectors}
\label{sec:new_data}
We conducted a set of experiments on larger datasets (approximately $10,000$ samples) of varying classes, where the feature vectors are generated using auto-encoders. For all the experiments, we used autoencoder dimensions of $\dim{(X)}-500-500-2000-10$. The training of the autoencoder included 50 epochs of layerwise training, followed by 50 additional epochs for fine-tuning the entire autoencoder.

In Table \ref{tab:autoencoder}, we report a full quantitative comparison on the YouTube Faces (YTF) dataset \cite{wolf2011face} which contains $10,000$ samples of faces, on the USPS dataset which contains $11,000$ samples of handwritten digits, and on the REUTERS dataset which contains $10,000$ samples of English news stories \cite{lewis2004rcv1}. As the table illustrates, our method obtains competitive performance on all three datasets, \frev{also in comparison to parametric methods that are provided with the number of clusters in the data}. Furthermore, our method does not over-segment these large datasets (unlike some of the other methods). In the supplementary material, we demonstrate that our method is insensitive to small variations of its automatically estimated parameter by reporting our scores on these datasets over a wide range of different parameter values. 

\subsubsection{Experiments on supervised feature vectors}
To further validate the performance of our technique on large datasets, we generated large sets by extracting feature vectors generated by convolutional neural networks (CNN) that were trained separately on MNIST \cite{lecun-mnisthandwrittendigit-2010}. MNIST is a well-known handwritten digit image dataset which consists of 70000 labeled images of handwritten digits divided into a training set of 60,000 images and a test set of 10,000 images.

\begin{figure*}[!t]
	\centering
	\vspace{-5pt}
	 \includegraphics[width=\textwidth]{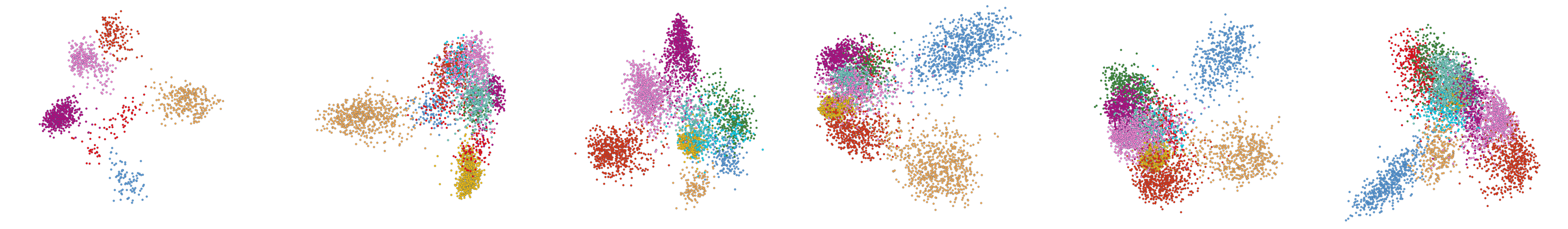}\\
\vspace{-10pt}
	\caption{Embeddings in 2d for selected samples of the MNIST features that were produced by a CNN. The embeddings were obtained by running PCA on the datasets.
}
\label{fig:mnist_cifar}
\end{figure*}

To obtain a strong feature representation on the MNIST samples, we trained a CNN using the training set and then used it to produce an embedding of the images in the test set to $n$-dimensional feature vectors ($n=500$). To obtain the feature vectors, we used the CNN implementation which is available in the MatConvNet \cite{vedaldi15matconvnet} library.
To produce varying less balanced datasets with an unknown number of clusters, the embedded test sets were sampled by taking all the embedded vectors which are within a certain radius of randomly sampled points. By using different radii, we generated several datasets of varying sizes. \frev{The selected radii values yielded varying datasets containing roughly a few thousand elements, spanning most, but not necessarily all, the digits (as illustrated on the top of Table \ref{tab:mnist}). On average, the number of sampled images is 2151, 3115, 4321 and 4471 for radius size 120, 130, 140 and 150, respectively.} Embeddings of selected samples of the MNIST dataset to 2D can be seen in Figure \ref{fig:mnist_cifar}. As a final preprocessing step, we employed PCA on the sampled datasets to avoid clustering in an overly-high dimensional space, reducing the original dimensions down to $30$.

\begin{table}
\centering
\ra{0.8}
\setlength{\tabcolsep}{2.05pt}
\begin{tabular}{@{}llrrrrrrrrrrrrrrrrrrrrrrrrrr@{}}
\toprule
\multicolumn{2}{c}{} &  & \multicolumn{3}{c}{YTF (41 Classes)} & \phantom{}& \multicolumn{3}{c}{USPS (10 Classes)} & \phantom{}& \multicolumn{3}{c}{REUTERS (4 Classes)}  \\
 && & ARI & AMI & Det\# &&  ARI & AMI & Det\# &&  ARI & AMI & Det\# \\ \midrule
 \frev{KM} && & 0.587&0.760 & --  && 0.517 & 0.616 & -- && \textbf{0.574} &\textbf{0.494} & -- \\  
\frev{ SC} && & 0.203&0.569 & --   && 0.533 & 0.714 &-- &&  0.005&	0.011 & --\\ \midrule
  DB && & 0.016 &0.220 & 312  && 0.000 & 0.005 & 21 && 0.011 &0.021 & 64\\  
 HDB && & \textbf{0.657} & \textbf{0.765} & 111   && 0.088 & 0.285 &14 && 0.083 &0.063 & 3\\ 
 AP && & 0.308 &0.628 & 242   &&  0.053 & 0.347 &346 && 0.013 &0.178 & 409\\ 
 MS && & 	0.016 & 0.115 & 5437   && 0.000 & 0.003 & 8315 && 0.002 &0.003 & 7432\\  
 RCC && &0.435 & 0.645 & 464  &&\textbf{ 0.749} & \textbf{0.727} & 109 && \textbf{0.200} &0.243 & 329\\ 
 BP && & 0.252 &0.683 & 106   && 0.672 &  0.723 & 19 && 0.145 & \textbf{0.246} & 47\\ 
\bottomrule
\end{tabular}
	\caption{\frev{Comparison of Border-Peeling clustering (BP) with parametric (KM, SC) and non-parametric (HDB, DB, AP, MS, RCC) clustering techniques on large datasets in an unsupervised setting. Alongside the scores, we also report the total number of output clusters (Det\#).}
}
\label{tab:autoencoder}
\end{table}


\begin{table*}
\centering
\vspace{-5pt}
\ra{0.8}
\setlength{\tabcolsep}{2.05pt}
\begin{tabular}{@{}lcccccccccccccccccccccccccccccccccccc@{}}
\toprule
\multicolumn{2}{c}{} &  & \multicolumn{3}{c}{$full$} & \phantom{}& \multicolumn{3}{c}{$r=120$ (8.0 Classes)} & \phantom{}& \multicolumn{3}{c}{$r=130$ (8.7 Classes)} & \phantom{} & \multicolumn{3}{c}{$r=140$ (9.5 Classes)}  & \phantom{} & \multicolumn{3}{c}{$r=150$ (9.9 Classes)} \\
 && & ARI & AMI & \frev{Det\#} &&  ARI & AMI & \frev{Det\#} &&  ARI & AMI & \frev{Det\#} &&  ARI & AMI & \frev{Det\#} &&  ARI & AMI & \frev{Det\#} \\ \midrule
  DB && & 0.232 &0.438 & 24 && 0.869   & 0.845 & 8.6&& 0.667 &0.729& 9.0 &&  0.534 & 0.634& 9.0 &&  0.409 & 0.573& 8.4\\  
 HDB && & 0.204 &  0.505 & 20 && 0.934  & 0.899& 8.6  && 0.759 & 0.759& 8.7 && 0.730 & 0.792 & 9.9 &&  0.656 & 0.767  & 9.6\\ 
 AP && & 0.050 &0.354  & 405 && 0.265  &0.484 & 33.6 && 0.197 & 0.442& 57.0 && 0.184 & 0.443 & 75.6 &&  0.149 & 0.427 & 110.9\\ 
 MS && & 0.238 & 0.188  & 32 && 0.625  &0.714 & 7.9 && 0.544 & 0.693 & 8.4 &&  0.430 & 0.650 & 8.8 &&  0.333 &0.609 & 9.6\\  
 QCC && & 0.153 &0.233 & 9 && 0.131   & 0.098 & 53.7 && 0.142 & 0.176 &82.6 &&  0.247 & 0.263 & 94.7&&  0.229 & 0.226&115.5\\  
 RCC && & \textbf{0.936} & \textbf{0.932}  & 18 && \textbf{0.976}  & \textbf{0.929} &8.2&& \textbf{0.952} & \textbf{0.922} &8.6&& 0.796 & 0.817&15.2 &&  0.829 & 0.873 &18.5\\ 
 BP && & 0.855 & 0.836  & 14 && 0.932 &0.896 &7.9 &&  0.946 & 0.909& 8.9 &&  \textbf{0.928} & \textbf{0.890} & 10.0&&  \textbf{0.927} & \textbf{0.895} &10.6\\ 

\bottomrule
\end{tabular}
\vspace{-5pt}
	\caption{A comparison to non-parametric clustering techniques on samples of the MNIST dataset. For each radius size, we sample multiple dataset configurations, \frev{reporting the average metrics (ARI, AMI) and the average number of clusters (Det\#)}. In the leftmost column, we report the scores on the full datasets, without performing a random sampling strategy.
}
\label{tab:mnist}
\end{table*}

The results of running Border-Peeling clustering as well as the other non-parametric clustering techniques are illustrated in Table \ref{tab:mnist}. For each radii value, we perform ten different runs and average the results of the clustering methods over these random runs. As a baseline, we also report the results on the full (not-sampled) datasets.

\begin{figure}[t!]
	\centering
	\includegraphics[width=\columnwidth]{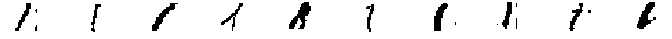} \\ \vspace{3pt}
	\includegraphics[width=\columnwidth]{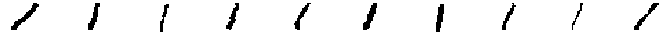} \\ \vspace{3pt}
	\includegraphics[width=\columnwidth]{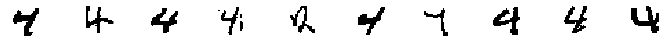} \\ \vspace{3pt}
	\includegraphics[width=\columnwidth]{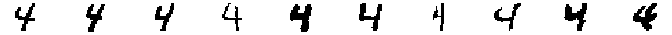} \\ \vspace{3pt}
	\includegraphics[width=\columnwidth]{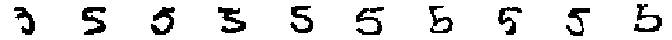} \\ \vspace{3pt}
	\includegraphics[width=\columnwidth]{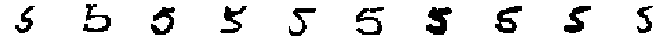} \\ \vspace{3pt}
	\includegraphics[width=\columnwidth]{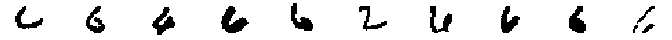} \\ \vspace{3pt}
	\includegraphics[width=\columnwidth]{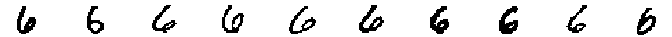} 
	\caption{The images corresponding to the bottom-10 (odd rows) and top-10 values (even rows) of $b_i^{(0)}$ for four different clusters which were obtained using Border-Peeling clustering on a random subset of the MNIST dataset.}
	\label{fig:mnist_confidence}
\end{figure}

As the table demonstrates, RCC and our technique achieve the best performance on average in terms of AMI and ARI, and for larger radii values our method outperforms the RCC technique. \frev{It is especially interesting to observe the huge variability in performance in this experiment where the features are obtained in a supervised fashion and indeed, for some methods, the scores are rather high in most cases. However, it seems that some common clustering solutions do not scale up to the large dataset size or the high dimensional feature space of the real image samples, even when provided with the supervised feature vectors. Furthermore, these experiments also demonstrate that our proposed method avoids over-partitioning to some degree, in particular in comparison to other methods.    }
Our incremental peeling process can also infer a confidence value associated with the data points of each cluster. 
As discussed in Section \ref{sec:BorderIden}, data points with lower values of $b_i$ are expected to be along the border of the cluster, and thus, with a lower confidence.
To illustrate this simple confidence ranking, Figure \ref{fig:mnist_confidence} visualizes the top-10 and bottom-10 instances of clustering the MNIST dataset.  Note that the images of digits with lower values of $b_i^{(0)}$ are often harder to identify, while images with higher values are clearer and more pronounced.

\subsection{Implementation details and runtime} 

The Border-Peeling clustering method was implemented in the Python programming language using the numpy \cite{van2011numpy} software library. In order to compare our performance with other clustering algorithms we used the implementation available in the SciKit-Learn \cite{scikit-learn} Python library. The time complexity of our technique is $O(T\cdot ( k \cdot n + \tilde{f}_{knn}))$, where $\tilde{f}_{knn}$ is the asymptotic complexity of the $k-NN$ method with respect to the dataset. 

\rev{
The runtime performance on a 2.5 GHz Intel Core i7 Mac-Book Pro on the datasets evaluated in Section \ref{sec:new_data} is as follows: $71$ seconds on the YTF dataset ($T=19$), $98$ seconds on the USPS dataset ($T=36$), and $106$ seconds on the REUTERS dataset ($T=23$). For comparison, the average runtime of a single run of the RCC technique on these datasets is $36$ seconds, $65$ seconds, and $55$ seconds, respectively.   
}
The implementation of our method is available at: \url{https://github.com/nadavbar/BorderPeelingClustering}.

\section{Conclusions}

We have presented a non-parametric clustering technique that groups points into different clusters by iteratively identifying points that reside on the borders of the cluster and removing them until separable areas of data remain. During the peeling process, the method creates associations between the peeled border points and points in the inner layers by estimating the local density 
of the data points. These associations are then used to link between the separable unpeeled data points and thus form the resulting clusters. 

The main idea of the method is the peeling of the border points which ensures that the cores of near-by clusters are clearly separated before the data points are classified as separate clusters. We present a novel formulation for identifying border points, which we validated analytically on a simple setting of a uniformly-distributed cluster. Additionally, as we have shown above, our border peeling technique can be used to infer confidence values associated with the data points of each cluster.
Unlike other methods, we do not make strong assumptions about the structure of the data points or their density distribution such as a single density peak or uniform density levels. 
Our approach can cope with multiple distribution models, as $k-NN$ neighborhoods, unlike $\epsilon$-neighborhoods, are not affected by varying distributions in the data.
Furthermore, the method has been shown to be stable in the sense that it is insensitive to the setting of hard coded parameters.


We have extensively analyzed the performance of the Border-Peeling method over large datasets for which the number of clusters is unknown, and there is no prior knowledge about its general structure or density distribution.  As we have shown, Border-Peeling clustering is competitive to other state-of-the-art non-parametric clustering methods, even when their free parameters are fine-tuned to achieve their best performance. In particular, the performance of Border-Peeling in comparison to DBSCAN is interesting since conceptually the two techniques have much in common, as both methods extract the core of the clusters that separate adjacent clusters, and then expand core points to the rest of the cluster. However, the difference in performance is intriguing. We attribute the better performance of Border-Peeling clustering to the fact that the core points are not defined globally, but through an iterative process that senses the local densities. 
The incremental peeling not only identifies the border points but also carefully associates them with points that seem to be closer to the core of the cluster. Furthermore, the locally adaptive approach is advantageous in sensing and avoiding over-segmentation of the clusters. Having said that, we believe that the more significant advantage of Border-Peeling over DBSCAN and other non-parametric clustering methods is that the method is insensitive to small variations in parameter values, and that those values can be easily set according to the characteristics of the data set, as can be seen in the experiments section. 

All our results were generated with the same set of parameters, which is detailed in the paper. To control the maximum association distance between data points, we estimated $\lambda$ (introduced in Section \ref{sec:BorderAssoc}) according to the global behavior of the data. We believe that this global estimation can be further refined to better suit datasets with different structures and densities, perhaps in a semi-supervised setting where some samples are labeled. In general, as future work, we believe that our unsupervised technique can be expanded to accommodate supervised scenarios, to achieve improved accuracy in such controlled settings.




\ifCLASSOPTIONcompsoc
\else
\fi


\ifCLASSOPTIONcaptionsoff
  \newpage
\fi


\bibliographystyle{IEEEtran}
\bibliography{bp_bib}

%

\vspace{-290pt}

\begin{IEEEbiography}[{ \vspace{-10pt} \includegraphics[width=0.95in,height=1.05in,clip, keepaspectratio]{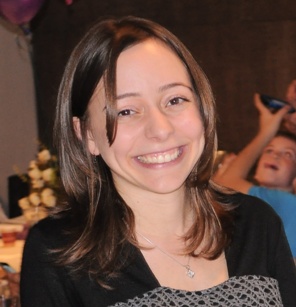}}]{Hadar Averbuch-Elor}
is a PhD student at the School of Electrical Engineering, Tel-Aviv University and a research scientist at Amazon AI. She received the BSc (cum laude) degree in electrical engineering from the Technion in 2012. She worked as an computer vision algorithms developer in the defense industry from 2011 to 2015. In 2016, she was a research intern in the Computational Photography group at Facebook. Her research interests include computer vision and computer graphics, focusing on unstructured image collections and unsupervised techniques.
\end{IEEEbiography}

\vspace{-290pt}

\begin{IEEEbiography}[{ \vspace{-19pt} \includegraphics[width=0.95in,height=1.05in,clip, keepaspectratio]{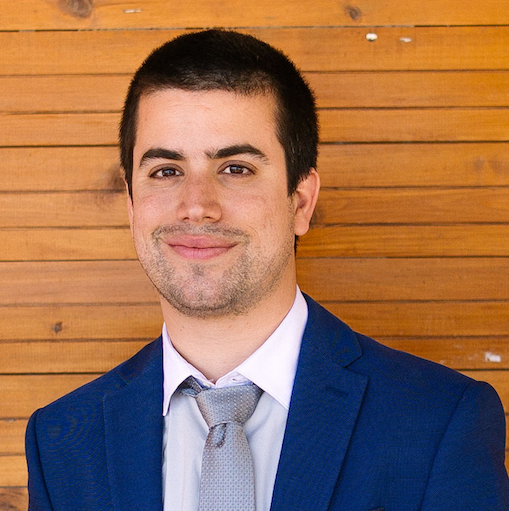}}]{Nadav Bar}
is a software engineer at Google. He received the BSc (cum laude) degree in computer science and computational biology from the Hebrew University of Jerusalem in 2012 and the MSc degree in computer science from Tel-Aviv University in 2017. His research interests include machine learning and, in particular, unsupervised learning and data clustering methods.
\end{IEEEbiography}

\vspace{-310pt}

\begin{IEEEbiography}[{ \vspace{-10pt} \includegraphics[width=1.05in,height=1.25in,clip, keepaspectratio]{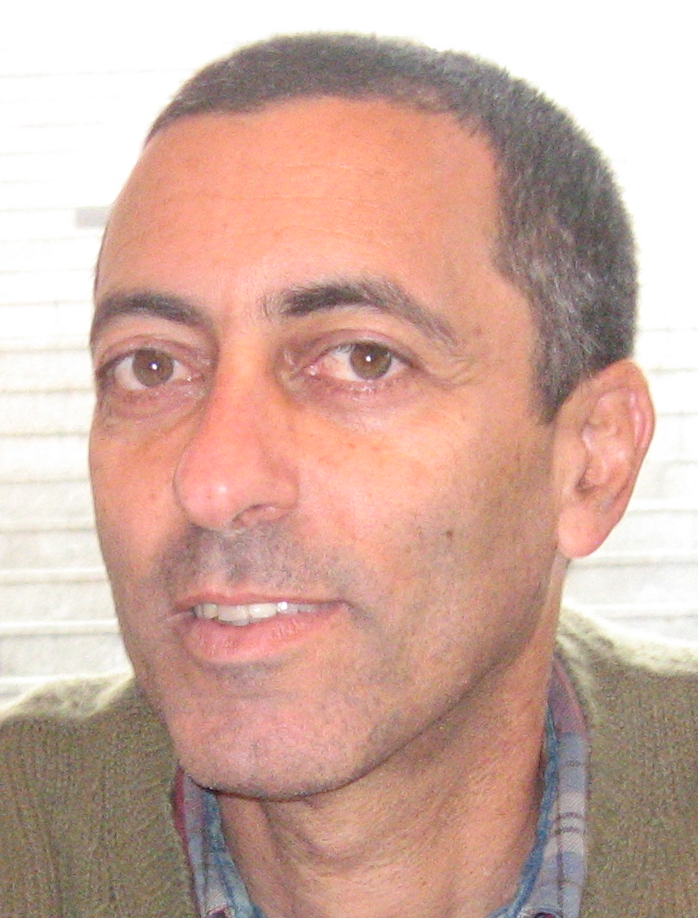}}]{Daniel Cohen-Or}
is a professor at the School of Computer Science, Tel-Aviv University. He received the BSc (cum laude) degree in mathematics and computer Science and the MSc (cum laude) degree in computer science, both from Ben-Gurion University, in 1985 and 1986, respectively. He received the PhD from the Department of Computer Science at State University of New York at Stony Brook in 1991. He received the 2005 Eurographics Outstanding Technical Contributions Award. In 2015, he was named a Thomson Reuters Highly Cited Researcher. In 2019, he won The Kadar Family Award for Outstanding Research. Currently, his main interests are in few areas: image synthesis, analysis and reconstruction, motion and transformations, shapes and surfaces.
\end{IEEEbiography}




\end{document}